\documentclass{IEEEtran}

\usepackage[utf8]{inputenc} 
\usepackage[T1]{fontenc}    
\usepackage{hyperref}       
\usepackage{url}            
\usepackage{booktabs}       
\usepackage{amsfonts}       
\usepackage{nicefrac}       
\usepackage{microtype}      
\usepackage{xcolor}         
\usepackage{graphicx}
\usepackage{amsmath}
\usepackage{booktabs}
\usepackage{microtype}
\usepackage{graphicx}
\usepackage{subfigure}
\usepackage{booktabs} 
\usepackage{graphicx}
\usepackage{float}
\usepackage{booktabs}
\usepackage{multirow}
\usepackage{siunitx}

\usepackage{hyperref}

\usepackage{amsmath}
\usepackage{amssymb}
\usepackage{mathtools}
\usepackage{amsthm}

\theoremstyle{plain}

\theoremstyle{definition}

\theoremstyle{remark}

\usepackage[utf8]{inputenc} 
\usepackage[T1]{fontenc}
\usepackage[british]{babel}
\usepackage{csquotes}
\usepackage{comment}
\usepackage[style=apa]{biblatex}
\usepackage{caption}
\DeclareLanguageMapping{british}{british-apa}
\title{Robust Representation Learning for Privacy-Preserving Machine Learning: A
Multi-Objective Autoencoder Approach}
\author{Sofiane Ouaari$^{1, 2}$, Ali Burak Ünal$^{1, 2, 3}$, Mete Akgün$^{1, 2, 3}$ and Nico Pfeifer$^{1, 2}$ \\
    $^1$Methods in Medical Informatics, Department of Computer Science, University of Tuebingen, Germany\\
    $^2$Institute for Bioinformatics and Medical Informatics (IBMI), University of Tuebingen, Germany\\
    $^3$Medical Data Privacy and Privacy Preserving Machine Learning, University of Tuebingen, Germany\\

    \{sofiane.ouaari, ali-burak.uenal, mete.akguen, nico.pfeifer\}@uni-tuebingen.de}
\begin{document}
\maketitle
\footnote{Preprint. Under review}
\begin{abstract}
Several domains increasingly rely on machine learning in their applications. The resulting heavy dependence on
data has led to the emergence of various laws and regulations
around data ethics and privacy and growing awareness of the
need for privacy-preserving machine learning (ppML). Current
ppML techniques utilize methods that are either purely based on
cryptography, such as homomorphic encryption, or that introduce
noise into the input, such as differential privacy. The main
criticism given to those techniques is the fact that they either are
too slow or they trade off a model’s performance for improved
confidentiality. To address this performance reduction, we aim to
leverage robust representation learning as a way of encoding our
data while optimising the privacy-utility trade-off. Our method
centers on training autoencoders in a multi-objective manner
and then concatenating the latent and learned features from the
encoding part as the encoded form of our data. Such a deep
learning-powered encoding can then safely be sent to a third
party for intensive training and hyperparameter tuning. With our
proposed framework, we can share our data and use third party
tools without being under the threat of revealing its original form.
We empirically validate our results on unimodal and multimodal
settings, the latter following a vertical splitting system and show improved performance over state-of-the-art.
\end{abstract}

\section{Introduction}
A wide range of application sectors is drastically integrating machine learning (ML) in diverse products. A successful ML model often requires a huge amount of training data and powerful computational resources. However, the need for such enormous volumes of data to develop performing models raises serious privacy concerns. Such ML models might face multiple types of adversarial attacks depending on the type of access an adversary might have to the model (white or black-box). A membership inference attack \parencite{shokri2017membership} allows an attacker to query a trained machine learning model to predict whether a given example is in the model's training data set. On the other hand, an inversion attack \parencite{fredrikson2015model,wang2021variational,ye2022label} aims to recreate an input data point given a confidence score obtained from a black-box inference of the model. In order to make researchers and engineers take such privacy threats into consideration, many regulations and ethical data policies, such as GDPR, CCPA, and CPRA \parencite{hijmans2018ethical, rochel2021ethics} were set to raise awareness around this topic and restrict any data violations that might occur in a given ML pipeline. \\
Previous works have been done to reduce the effectiveness of different privacy attacks. Among those studies, differential privacy (DP) is the most commonly used approach which operates by incorporating predetermined randomization into a machine learning algorithm's computation. The perturbation introduced by DP might be applied on the users's input, parameters, prediction output and even on loss functions \parencite{abadi2016deep, phan2016differential}. However, many studies have shown that such noise reduces the performance of the model for the sake of privacy \parencite{truex2019hybrid}. Furthermore, Setting up a correct value for $\epsilon$ is complex by nature and requires some trial and error process.\\ 
Homomorphic encryption (HE) is a cryptographic method applied in the domain of ppML. It is defined as a type of encryption method that allows computations to be performed on encrypted data without first decrypting it with a secret key. Yet HE has some limitations. It was originally designed to allow only algebraic operations such as addition and multiplication which excludes the various non-linear activation functions leveraged in neural networks. Numerous studies have been conducted to approximate such functions using polynomials \parencite{hesamifard2017cryptodl,lee2021precise, lee2022privacy}, however such approximations result in high computational burden and reduce the ability to apply various methods by extending the depth of deep learning models, since performing all that in a HE fashion significantly increases computation time.  \\
The aim behind this paper is to create a deep learning-oriented encoding strategy by training a supervised residual autoencoder and concatenating the features learned from the encoder part as the representation to be shared with third parties for further training and an extensive hyperparameters search. A framework, which trains an autoencoder and shares its latent space embedding with other parties for data sharing purposes was presented by \cite{maria2022privacy}. We consider such a framework as a baseline in our experiments and empirically demonstrate (see Section \ref{exp_setup}) that our suggested architecture considerably improves the performance. In contrast to \cite{maria2022privacy}, we also provide a threat analysis to discuss how secure our framework is and the different types of access that an adversary might have, considering different actors directly involved and interacting with our framework. The autoencoder proposed in our framework is trained in a multi-objective fashion by simultaneously considering the data reconstruction and the supervised learning problem, and ensuring an informative and discriminative representation. This ppML encoding is applied to the data part of the machine learning pipeline and we empirically demonstrate the efficiency of this method by first experimenting on unimodal settings using the MNIST, FashionMNIST, Leukemia, and Retinal OCT datasets. In addition, we further explored the capabilities offered by our ppML framework on multimodal data distributed in a vertical setting, where each modality is provided by a given data party. For this purpose, a TCGA multi-omics breast cancer dataset was leveraged. 
We summarize our main contributions as follows:
\begin{itemize}
    \item We developed a considerably improved version of the data sharing strategy through latent space embedding proposed by \cite{maria2022privacy} by increasing the performance on different prediction tasks and providing a detailed threat analysis. 
    \item We demonstrated the application flow of our proposed encoding framework for both unimodal and multimodal (vertically distributed) settings. 
    \item We empirically validated our approach to be utility-privacy efficient by comparing models trained on the original data against models trained on the generated encoded data and show that both perform equally. 
\end{itemize}
\section{Background \& Related Work}
In this section, we present previous works that have been performed in the sphere of autoencoders, representation learning and ppML methods enhanced with deep learning. 
\subsection{Autoencoders}
Autoencoders are a type of neural network originally implemented to solve the unsupervised task of data reconstruction. Formally, an autoencoder is defined with three main parts, which are an encoder \textit{E(.)}, a latent space representation $s$ and a decoder \textit{D(.)}. Given $x, \hat{x}\in \mathbb{R}^{d}$ and $s \in \mathbb{R}^{m}$ we have $s = E(x)$ and $\hat{x} = D(s)$ with $m << d$ and $\hat{x}$ defined as the reconstructed output of the original input $x$. Beyond its original purpose, the use of autoencoders were extended to other applications such as data generation with variational autoencoders \parencite{kingma2013auto}, anomaly detection \parencite{sakurada2014anomaly} and recommendation systems \parencite{ferreira2020recommendation, pan2020learning}. They were also leveraged for supervised learning purposes, \cite{le2018supervised} implemented a supervised autoencoder (SAE) where the latent space $s$ is linked to a classifier $f_{c}$ trained in parallel with the original data reconstruction problem and the overall loss function defined as follows:

\begin{small}
\begin{multline}
    L(x, y, \theta_{e}, \theta_{d}, \theta_{c}) = 
    \frac{1}{t}\sum_{i=1}^{t} L_{r}(x,D(E(x, \theta_{e}),\theta_{d})) \\ + L_{c}(y,f_{c}(E(x, \theta_{e}),\theta_{c}))
\end{multline}
\end{small}
With $\theta_{e}$, $\theta_{d}$ and $\theta_{c}$ being the parameters of the encoder, decoder and classifier respectively and $L_{r}(.,.)$, $L_{c}(.,.)$ defined as reconstruction and categorical cross entropy losses.
\subsection{Representation learning}
\label{representation_learning}
A good encoding demands an informative representation of the original data by reducing the dimension of the input without lowering the inter-dependencies and the important relations needed for a given ML model to perform efficiently in a given task. A reasonable-sized learnt representation might encompass a vast array of potential input configurations, because good representations are expressive.
\textcite{bengio2013representation} surveyed what makes the essence of a good representation which we took into consideration while developing our framework. 
\begin{itemize}
    \item \textit{Smoothness}: Given $x, y \in \mathbb{R}^d$ and a representation function $f(.)$ defined as: $f:\mathbb{R}^d \rightarrow \mathbb{R}^{m}$, where $m<d$, a smooth representation implies that if $x \approx y$ then $f(x) \approx f(y)$. In a geometric point of view using distances this translates to: if $dist(x_1, x_2) < dist(x_1, x_3)$ then $dist(f(x_1), f(x_2)) < dist(f(x_1), f(x_3))$. This is a reason why we introduced a center loss in our framework which is explained in Section \ref{section_modified_center_loss} to keep a similar distance semantics when encoding the data and mapping it to a space with reduced dimension. 
    \item \textit{Sparsity $\&$ Invariance}: for any given observation in $x$, only a small fraction of the possible factors are relevant. In terms of representation, this could be represented by features that are often zero or by the fact that most of the extracted features are insensitive to small variations of $x$. For our case, this can be achieved through sparse autoencoders \parencite{rangamani2018sparse} which is a sort of autoencoder that uses sparsity to create a bottleneck in the flow of information. In particular, the loss function is designed to punish the activation generated by the layer. $L1$ regularization is usually used to apply the sparsity constraint. 
\end{itemize}
\subsection{Privacy enhancement with deep learning}
Linking deep learning training concepts for the purpose of ppML has been addressed before. Adversarial learning is the most frequent way to handle privacy-utility trade-off where the privacy variable is explicitly introduced in the adversarial objective.  \textcite{mandal2022uncertainty} presented UAE-PUPET where an autoencoder takes an input $x$ and generates $\hat{x}$ being a distorted version of $x$ that contains minimum information of a given private attribute $x_p$ while always keeping the most important information of the targeted utility variable $x_u$. UAE-PUPET works by linking 2 classifiers to the autoencoder $\gamma_u$, $\gamma_p$ responsible on predicting the utility and private attributes respectively, then adversarialy minimizing $L_u$ and maximizing $L_p$ while added to the original reconstruction loss. However, such a method always requires choosing only one privacy variable at a time which is not always the case, in addition the authors did not run a clear threat analysis through analyzing how a harmful attack as model inversion might be applied against the generated $\hat{x}$.\\
\textcite{xiao2020adversarial} proposed an adversarial reconstruction learning framework that prevents the latent representations to decode into the original input data. In other words, this time the reconstruction loss is directly maximized while minimizing the utility prediction error. Even though such a method allowed to empirically reduce the effectiveness of a model inversion attack, it is theoretically still possible and not inevitable. In addition the authors also noted that adversaries may still be able to exploit private meta information, such as determining the ethnicity of an individual based on skin color presented in reconstructed images, despite the fact that the images are not identical to the input.\\
Training a model $\mathcal{M}$ with generated synthetic data is another ppML with deep learning approach. Previous works were performed to train Generative Adversarial Neural networks (GANs) using differential privacy DP-GAN \parencite{wang2021dpgen, cao2021don, harder2021dp}. Yet, DP leads to the injection of noise in the generated data, which might reduce the model's performance especially in complex data like genomics where the data suffers from the curse of dimensionality and where one single gene can decide the main task's outcome, DP might result in a further performance loss \parencite{chen2020differential}.\\
\textcite{dong2022privacy} were the first to introduce dataset condensation technique into the domain of ppML. Data condensation \parencite{zhao2021dataset} works by generating synthetic data through condensing the larger original data points into a smaller subset. It allows a data-efficient learning since the synthetic data are created by first randomly initializing them, afterwards two neural networks are trained where one is linked to the original data $\mathcal{T}$ and the other to the synthetic data $\mathcal{S}$ and then the latter is updated iteratively by back-propagating the loss in a way that both networks share the same weights. \textcite{dong2022privacy} theoretically proved that dataset condensation is similar to differential privacy from the perspective that one sample has limited effect over the parameter distribution of a network. Then they empirically showed that in addition to keeping good performance a model trained with the synthetic data was also more robust towards membership inference attacks. However, such a method does present some limitations. First, the membership inference risk was still present even though it was decreased, in fact the authors even found out that in the case of FashionMNIST the membership attack was more
effective on the model trained with synthetic data $\mathcal{S}$ than on its original counterpart due to the grey-scale nature of the image, assuming that the synthetic data might contain more features prone to be memorized. Furthermore, dataset condensation initially requires a large amount of data, which excludes databases with few samples.
\section{Framework Architecture \& Methodology}
\subsection{Residual Autoencoder}
\label{residual_autoencoder_section}
\begin{figure}[ht]
\begin{center}
\centerline{\includegraphics[height=3cm,width=2cm, angle=90]{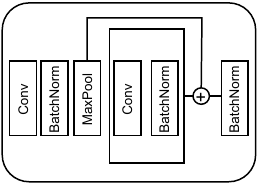}}
\caption{Building block of the Convolution Residual Autoencoder (C-RAE) in the encoder part $E(.)$ }
\label{backbone-architecture}
\end{center}
\end{figure}

Ensuring the training stability in a neural network raises several questions, especially on how to correctly set the number of hidden layers to be used. \textcite{wickramasinghe2021resnet} showed that residual autoencoders (RAE) performed well and in a stable manner while ranging the number of repeated layers from 2 to 90 on MNIST, FashionMNIST and CIFAR. For this stability purpose, we decided to use the residual autoencoder as the main backbone of our encoding framework. By extension, we used a Convolution-RAE (C-RAE) when encoding images (Fig \ref{backbone-architecture}). Then,
every layer in the encoder part including the latent space is forwarded in parallel to a classifier.
Formally, each hidden layer of the encoder $X_{enc}^{(i)} \in \mathbb{R}^{e_{i}}$, with $e_{l}<e_{i+1}< e_{i}<d$ and $d, e_{l}$ being the respective dimensions of the input data and the latent space ($e_{l} < d$), is separately fed into a neural network classifier $f_{i}(\cdot)$ to perform the original classification task assigned to it $\hat{y}_{i} = f_{i}(X_{enc}^{(i)},\theta_{i})$, $\hat{y}_{i} \in C$ and $i$ is the index of layers included in the encoder $E(.)$ with the latent space taking part as well.  

\subsection{Application Flow}
In this section, we explain the application pipeline of our encoding framework in both unimodal and multimodal schemes.
\subsubsection{Unimodal Setting}
\begin{figure}[ht]
\begin{center}
\centerline{\includegraphics[width=6cm]{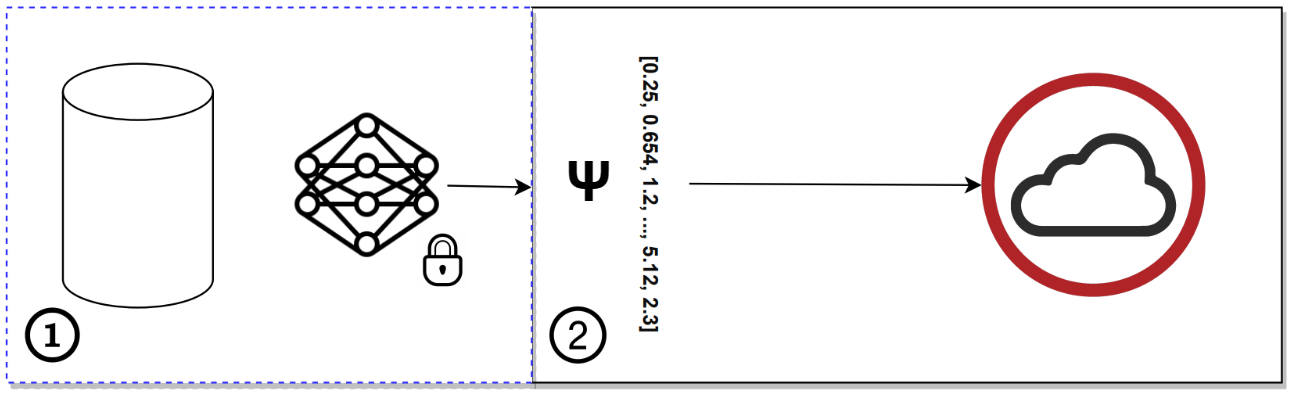}}
\caption{Execution flow of our ppML encoding framework on unimodal data, in $\textcircled{\tiny{\textbf{1}}}$ our supervised residual autoencoder is trained in-house on the original data, then in $\textcircled{\tiny{\textbf {2}}}$ the generated concatenated encoding \scalebox{1.2}{$\Psi$}  will be sent and shared with another party (for example: cloud) for the purpose of performing a heavy hyper-parameters tuning and further computationally expensive experiments.}
\label{unimodal-encoding-setup}
\end{center}
\end{figure}

After training the supervised residual autoencoder (SRAE) of our framework on a dataset $\{x_j,y_j\}$, we use the encoder part $E(.)$ of it to compute different representation of a given sample $x_j$ until the layer producing the latent space as defined in the previous section. The final encoding \scalebox{1.2}{$\Psi$} to be shared is defined simply as the concatenation of those representations (Eq. \ref{xi_final}). 
\begin{small}
\begin{equation}
    \scalebox{1.2}{$\Psi$} = (X^1_{enc}, X^2_{enc},..., X^l_{enc})
    \label{xi_final}
\end{equation} 
\end{small}
with $X_{enc}^{(i)} \in \mathbb{R}^{e_{i}}$ and $\scalebox{1.2}{$\Psi$}  \in \mathbb{R}^{\sum_{i=1}^{l}e_{i}}$. In Fig \ref{unimodal-encoding-setup}, we illustrate how the encoding pipeline is generated by our proposed framework for a unimodal scenario. 
\\
We would like to highlight and point out that in the unimodal setting our encoding framework is helpful when further computational power from external resources is required, for instance to evaluate different hyperparameters settings and combinations  and/or the use of complex ensemble and mixture of experts models. In other words, the trained residual autoencoder is kept in-house and only the encoded data is shared with the server providing the compute resources allowing us to benefit from cloud services without being under the threat of revealing our original data $x$. Our framework can also be leveraged to permit other institutions, when requested, to benefit from the inference of our model by sharing with them both the encoder $E(.)$ and the final model trained on the cloud $\mathcal{M}_\Psi$. 

\subsubsection{Multimodal Setting}
In this section of the paper we explain how to extend the utility of our framework in a multimodal use case. Let us consider a multimodal dataset consisting of $m$ modalities $\{x^{(m)}_{j},  y_{j}\}$ and distributed in a vertical setting over $m$ data providers where basically each modality is stored in a data supplier.
\begin{figure}[ht]
\begin{center}
\centerline{\includegraphics[width=5cm]{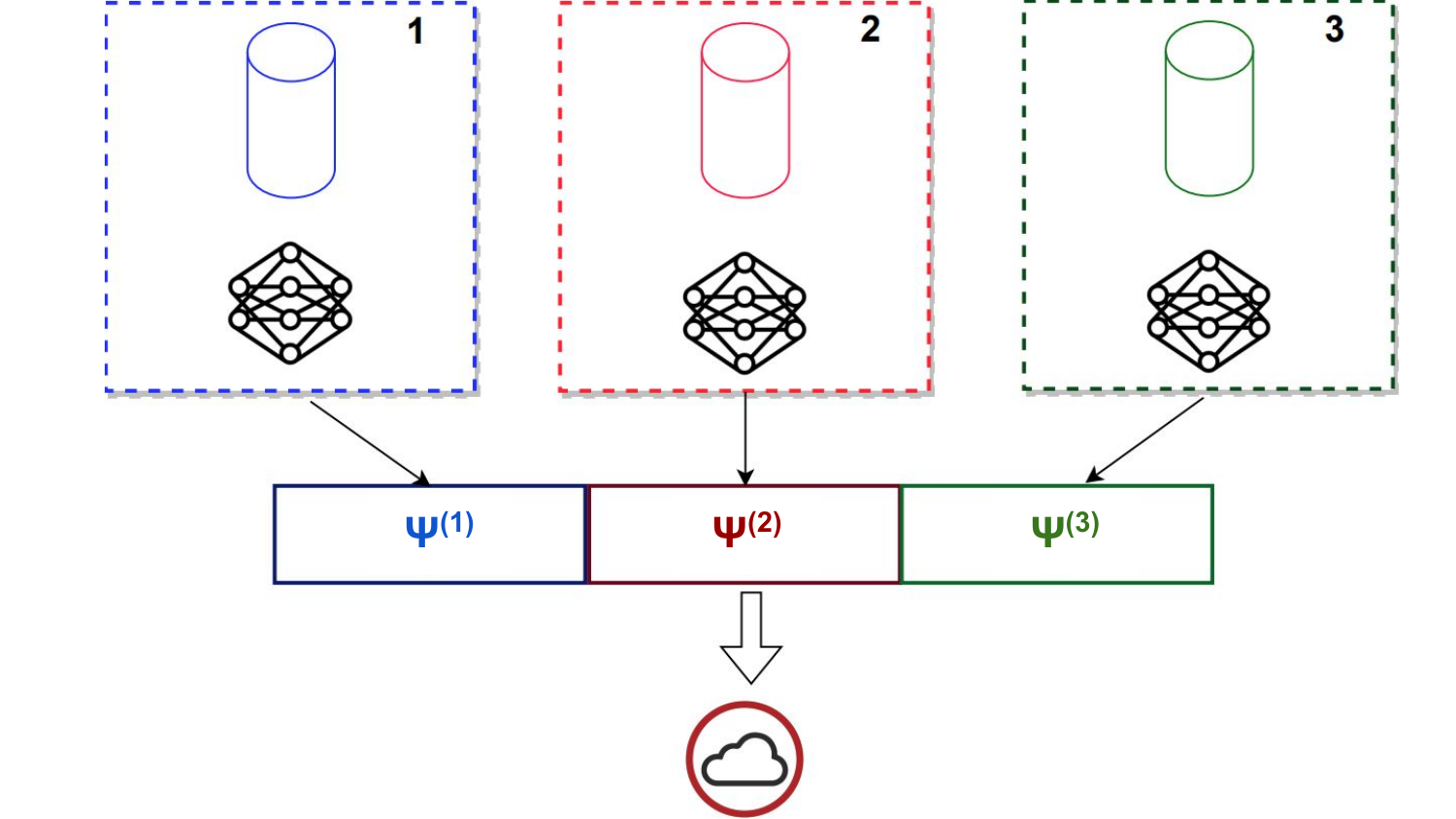}}
\caption{Execution flow when applying our ppML encoding framework on a multimodal use case with vertically distributed data}
\label{multimodal-encoding-setup}
\end{center}
\end{figure}
 For this specific scenario, the adopted strategy works by training a supervised residual autoencoder on every modality separately at the data provider. Each one then sends in parallel its final encoding $\scalebox{1.2}{$\Psi$} ^{(m)}$ to a third party system to perform a cooperative training (Fig \ref{multimodal-encoding-setup}). Such a workflow permits mutual work between various providers by sending good representations of the input thus not hurting the performance and in the same time guaranteeing total privacy of the data since no information about the original data format or properties are revealed including the original shape, the distribution, the type of modality (image, sound, tabular...) and if the data-type is homogeneous (only numerical, only categorical) or heterogeneous.\\
For the sake of a better understanding of how the proposed encoding framework can be applied on a multimodal scenario, we consider the following illustrative example. In the healthcare domain, 3 clinics are collaboratively working together to diagnose a given patient $P_j$ with a certain pathology $H_k$. Each clinic is responsible for delivering specific information about the patient $P_j$. For instance one presents the X-ray image, the second delivers the IRM image and the last one shares the electronic health record (EHR). Our framework allows the 3 institutions to collectively train a model responsible for detecting a pathology $H_k$ and to infer if patient $P_j$ suffers form it, all that without revealing the true format of the data to other parties in a way that clinic 2 and 3 are not able to know the original content of the X-ray image stored in clinic 1.    
\subsection{Multi-Objective Paradigm}
Our model, as previously mentioned, is trained to solve multiple tasks simultaneously. In addition to considering both the data reconstruction and classification problems in parallel, two other tasks within the sphere of representation learning are taken into account by the model.
\subsubsection{Center Loss}
\label{section_modified_center_loss}

In order to ensure that the layers' concatenation which will be shared by our framework has a widely class-separated structure to maximize the distance from decision boundaries we introduced a center loss on the concatenation layer. Center loss was presented by \textcite{wen2016discriminative} and defined in Eq. \ref{org_center_loss_eq}, where $c_{y_i} \in \mathbb{R}^{d}$ denotes the $y_{i}$-th class center. It aims to minimize the intra-class distance.
\begin{small}
    \begin{equation} 
    L_c =  \sum_{i=1}^{n}  ||x_{i} - c_{y_i}||_{2}^{2}
\label{org_center_loss_eq}
\end{equation}
\end{small}

\subsubsection{Cosine Similarity with PCA}

\begin{figure}[ht]
\captionsetup{justification=centering}
\begin{center}

\subfigure[Summary architecture of our proposed encoding framework]{
\includegraphics[width=7.5cm]{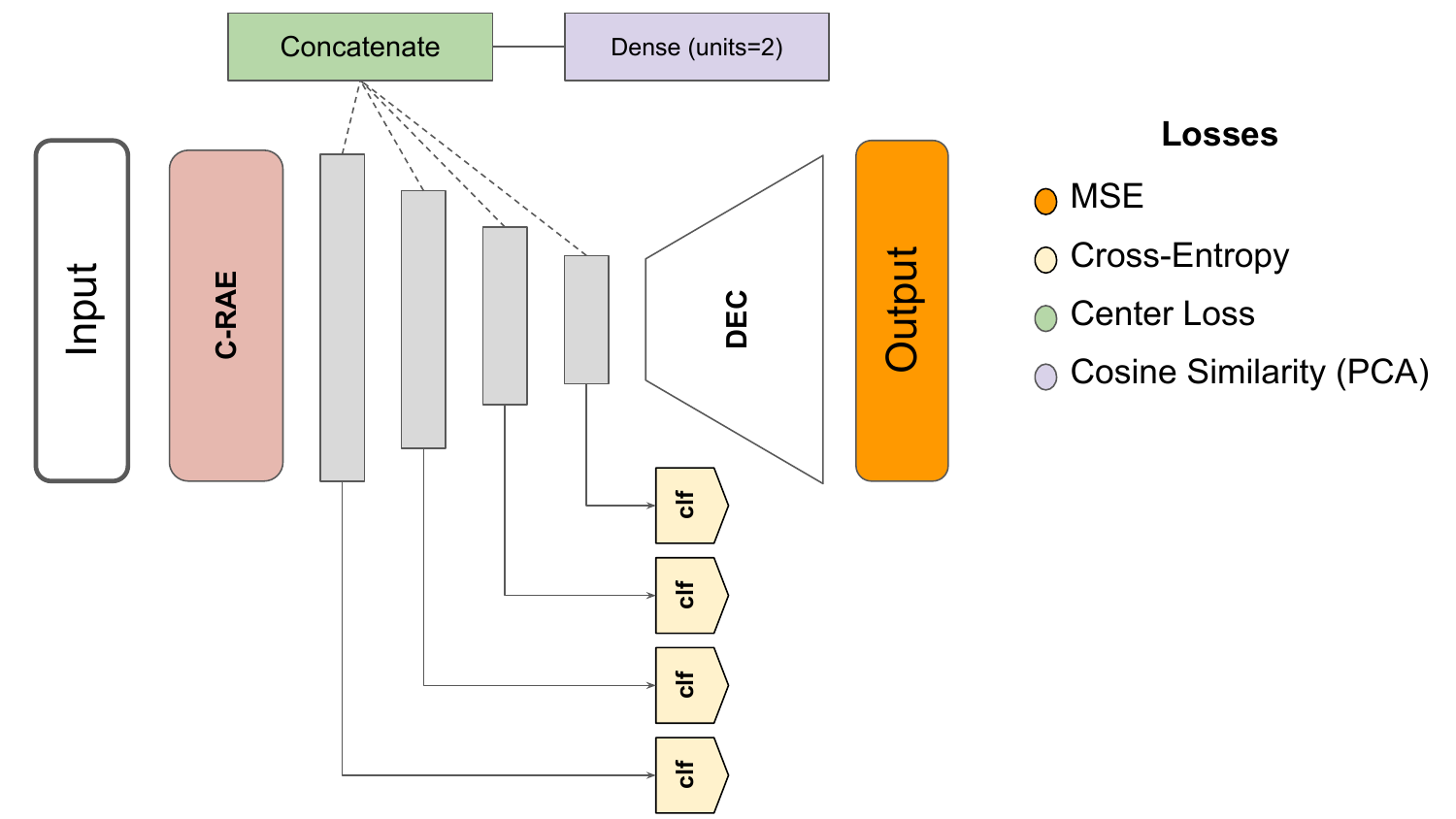}
}
\subfigure[Baseline architecture proposed by \cite{maria2022privacy}]{\includegraphics[width=5cm]{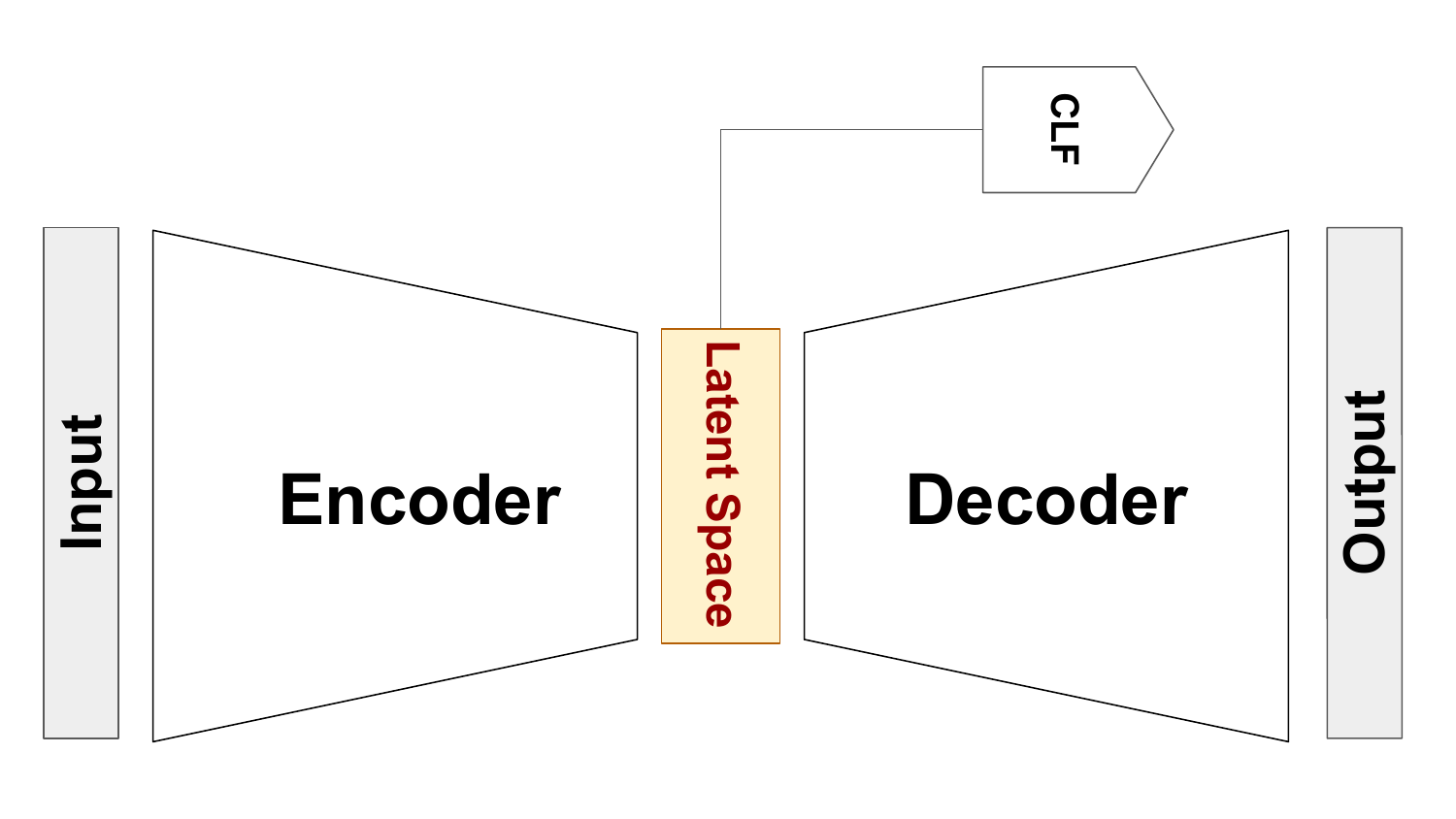}}
\caption{Suggested Framework vs Baseline}
\label{summary-architecture}
\end{center}
\end{figure}

Explicitly introducing an interpretation mechanism in our representation  is of crucial importance. Since the "black-box" barrier is always present when training neural network models such as autoencoders, we wanted to explicitly make the learning of the representation aligned with the PCA of the original data, as PCA is a well accepted technique for dimensionality reduction.\\
For this aim, we decided to use a cosine similarity loss function. In the same network, we connect the concatenation layer, where the center loss is already applied (Section \ref{section_modified_center_loss}), to a dense layer of 2 units and minimize the following function: 
\begin{small}
\begin{equation}
    L_{pca} = 1 - \frac{f_2(\scalebox{1.2}{$\Psi$} ) \cdot x_{pca}}{\|f_2(\scalebox{1.2}{$\Psi$} )\| \|x_{pca}\|}
\label{cosine_similarity_pca}
\end{equation}
 \end{small}
and $f_2(.)$ defined as $f_2:\mathbb{R}^{\sum_{i=1}^{l}e_{i}} \rightarrow \mathbb{R}^{2}$ and $x_{pca}$ being the 2-d PCA of the original input $x$. 
All parts of our framework are summarized in Fig \ref{summary-architecture}. 
\section{Datasets}
We will now define the list of datasets we experimented on to evaluate the efficiency  of our encoding framework in terms of optimizing the performance-privacy trade-off. We went beyond using only standard benchmark datasets by also including data characterized with real life constraints as data imbalance and limited number of data points with features exceeding the number of samples. 
\subsection{Unimodal Dataset}
\subsubsection{Image Data}

 \textbf{MNIST} \parencite{lecun1998gradient} is considered a benchmark for image classification and includes grey scale images of digits.
\textbf{Fashion MNIST} \parencite{xiao2017fashion} is a benchmark dataset for machine learning similar to MNIST, but is by nature more complex since the task is to identify 10 types of Zalando articles.
\textbf{Retinal Optical Coherence Tomography} (OCT) is an imaging technique used to capture high-resolution cross sections of the retinas. The dataset presented by \textcite{kermany2018identifying} is comprised of 84 495 retina images with 4 classes: Normal, Diabetic Macular Edema, Drusen and Choroidal Neovascularization. 

\subsubsection{Tabular Data}
\textbf{Leukemia}: this tabular dataset comes from the Curated Microarray Database \parencite{Feltes2019} which is a repository containing 78 handpicked cancer microarray datasets, extensively curated from 30.000 studies from the Gene Expression Omnibus (GEO), solely for machine learning.  For our study, we experimented on the leukemia dataset which contains 281 samples with 22284 gene expression values with the task of classifying 7 types of leukemic cancer.\subsection{Multimodal Dataset}\textbf{TCGA Breast Information Core} is a multi-omics dataset for breast cancer. Different types of high-throughput sequencing methods used parameters of DNA genome sequence, RNA expression and DNA methylation. Each datatype is labeled with the term ``omics'' (genomics, transcriptomics and methylomics respectively for our case). This dataset was leveraged by \textcite{rappoport2018multi}, \textcite{roder2019web} and \textcite{cantini2021benchmarking} to perform and experiment on multi-omics and multi-view clustering methods. However, since the current version of our representation-oriented encoding framework is specific to supervised learning use cases, we applied it to predict the survival status of patients.

\section{Experiments \& Evaluation}
In this section, we present experiments conducted to compare the performance of training models on the encoded data against models directly trained on the original data or trained only on the latent space embedding (\cite{maria2022privacy}). 
\subsection{Model Parameters}
Details about the type, number, and hyperparameters of the layers composing the residual autoencoder in our framework are presented in Appendix A. Hardware specifications are presented in Appendix B. 

\subsection{Experimental Setup}
\label{exp_setup}
For every dataset, we trained the residual autoencoder of our encoding framework on the train set, then obtained the concatenated encoding $\scalebox{1.2}{$\Psi$}^{train}$ and $\scalebox{1.2}{$\Psi$}^{test}$ for the train and test sets respectively. We then trained a set of machine learning models (KNN, SVM, Decision Trees, Random Forests, Multi-Layer Perceptron) in a randomized grid search manner (Appendix C) on the original data $x^{train}$, the latent space embedding only (baseline) and on our encoded data $\scalebox{1.2}{$\Psi$}^{train}$ and compared the performance metrics on the respective test points. However, for MNIST, FashionMNIST, and Retinal OCT since they are of type image we trained their original versions using a ResNet-50. 
 From table \ref{classification_results} we can empirically observe that the average macro F$_1$-score of models trained on the encoded data is most of the time better than the other approaches. 
 \begin{table*}[ht]
\caption{Macro F1-Score performance with Randomized Grid Search training in a 10-fold stratified cross-validation between using the original data (Org) vs Data shared as the latent space only, being the baseline, (BASE, \cite{maria2022privacy}) vs the encoded data outputted by our suggested ppML framework (Enc)}
\label{sample-table}
\begin{center}
\resizebox{\textwidth}{!}{
  \begin{tabular}{lSSSSSS}
    \toprule
    \multirow{3}{*}{} &
      \multicolumn{3}{c}{Leukemia} &
      \multicolumn{3}{c}{TCGA}  \\
      \bottomrule
      & {Org} & {Base} & {Enc} & {Org} & {Base} & {Enc} \\
      \midrule
    KNN & \emph{61.84 $\pm$5.18} & \emph{77.71 $\pm$7.57} & \textbf{82.57 $\pm$5.01} & \emph{60.11 $\pm$5.65} &  \emph{59.10 $\pm$4.56} & \textbf{60.77 $\pm$5.97} \\
    SVM & \textbf{83.79 $\pm$5.66} & \emph{78.81 $\pm$5.17} & \emph{80.52 $\pm$5.85} & \emph{33.96 $\pm$3.20} & \emph{59.62 $\pm$5.12} & \textbf{62.62 $\pm$3.61} \\
    DT & \emph{66.24 $\pm$3.11} & \emph{70.11 $\pm$4.35} & \textbf{74.97 $\pm$6.81} & \emph{56.97 $\pm$5.42} & \textbf{59.73 $\pm$5.23} & \emph{57.95 $\pm$5.15}\\
    RF & \emph{78.70 $\pm$2.80} & \emph{78.59 $\pm$ 5.91} & \textbf{83.69 $\pm$4.30} & \emph{61.96 $\pm$6.47} & \emph{59.64 $\pm$4.38} & \textbf{63.96 $\pm$4.16} \\
    MLP & \emph{82.06 $\pm$4.34} & \emph{80.36 $\pm$ 5.51} & \textbf{83.96 $\pm$3.17} & \emph{59.74 $\pm$4.64} & \emph{60.27 $\pm$ 5.96} & \textbf{62.79 $\pm$3.33} \\
    \bottomrule
  \end{tabular}}
\begin{tiny}
\resizebox{\textwidth}{!}{
  \begin{tabular}{lccccccccccccr} 
\\
\toprule
\textbf{Dataset} & \textbf{KNN-BASE} &\textbf{KNN-ENC} & \textbf{SVM-BASE} &\textbf{SVM-ENC} & \textbf{DT-BASE} &\textbf{DT-ENC} & \textbf{RF-BASE} &\textbf{RF-ENC} & \textbf{MLP-BASE} &\textbf{MLP-ENC} & \textbf{ResNet50-ORG}
\\ \midrule
MNIST & \emph{99.44 $\pm$ 0.05} & \emph{99.40 $\pm$0.02} & \textbf{99.47 $\pm$ 0.06} & \emph{99.44 $\pm$0.02} & \emph{99.16 $\pm$ 0.10} & \emph{99.29 $\pm$0.03} & \emph{99.44 $\pm$ 0.06} & \emph{99.43 $\pm$0.03} & \emph{99.44 $\pm$ 0.05} & \emph{99.41 $\pm$0.06} & \emph{99.31 $\pm$0.08}\\ \midrule
Fashion & \emph{92.34 $\pm$ 0.17} & \emph{92.29 $\pm$0.22} & \emph{92.25 $\pm$ 0.16} & \textbf{92.58 $\pm$0.22} & \emph{90.95 $\pm$ 0.43} & \emph{91.67 $\pm$0.41} & \emph{92.34 $\pm$ 0.16} & \emph{92.44 $\pm$0.21} & \emph{92.40 $\pm$ 0.16} & \emph{92.13 $\pm$0.18} & \emph{91.21 $\pm$0.33}  \\ \midrule
OCT & \emph{98.33 $\pm$ 1.45} & \emph{98.43 $\pm$0.26} & \emph{97.68 $\pm$ 1.91} & \emph{98.56 $\pm$0.22} & \emph{97.68 $\pm$ 1.91} & \emph{98.33 $\pm$0.28} & \emph{99.02 $\pm$ 0.45} & \textbf{99.19 $\pm$0.19} & \emph{98.20 $\pm$ 1.42} & \emph{98.58 $\pm$0.14}  & \emph{97.01 $\pm$1.28}\\
\bottomrule
\end{tabular}}

\end{tiny}
\end{center}
\label{classification_results}
\end{table*} 
We further explored the impact of the introduced center loss by comparing the silhouette score of the concatenated encoding \scalebox{1.2}{$\Psi$} with and without applying center loss. Prior to calculating the silhouette score we first reduce the concatenated encoding to 2 dimensions using t-SNE. 
As a reminder, silhouette score is a quantitative metric which measures how good clusters are grouped together. 

Table \ref{silhouette_score_results} summarizes the silhouette score comparison between two versions of our framework (with/without center loss). We observe an increase of the silhouette score by 478\%, 43\%, and 57\% for FashionMNIST, Leukemia, and TCGA, respectively after adding the center loss and a small decrease of 3\% and 0.5\% for MNIST and OCT. We clearly notice the positive impact of the center loss in making the encoded representation well grouped. 

We also applied an ablation study to check the impact of the introduced cosine similarity loss (Eq. \ref{cosine_similarity_pca}). Even though the main reason of introducing this loss was to properly guide the alignment of the input and the shared representation, we also evaluated its impact on the overall classification performance. For this case, we only took the best type of ML model per dataset while using the encoded data in Table \ref{classification_results}. The results of this ablation study are presented in Table \ref{pca_cosine_similarity_loss}.
\begin{table}[t]
\caption{Comparison between the silhouette score applied on the final concatenated encoding of the test set \textbf{$\scalebox{1.2}{$\Psi$}^{test}$} for 10-fold stratified cross-validation to evaluate the effect of adding center loss}
\begin{center}
\begin{tiny}
\begin{tabular}{lcccccr} 
\toprule
\textbf{$L_c$} & \textbf{MNIST} & \textbf{Fashion} & \textbf{Leukemia} & \textbf{OCT} & \textbf{TCGA}
\\ \midrule
Without & \textbf{0.597}  & \emph{0.078} & \emph{0.095} & \textbf{0.734} & \emph{0.048} \\ \midrule
With & \emph{0.594}  & \textbf{0.451} & \textbf{0.136} & \emph{0.712} & \textbf{0.076} \\ 
\bottomrule
\end{tabular}
\end{tiny}
\end{center}
\label{silhouette_score_results}
\end{table}

\begin{table}[t]
\caption{For each dataset, we took the best model found in Table \ref{classification_results} based on our encoding framework and evaluated the effect of the PCA cosine similarity loss.}
\begin{center}
\begin{tiny}
\begin{tabular}{lcccccr} 
\toprule
\textbf{$L_{pca}$} & \textbf{MNIST$_{SVM}$} & \textbf{Fashion$_{SVM}$} & \textbf{Leukemia$_{MLP}$} & \textbf{OCT$_{RF}$} & \textbf{TCGA$_{RF}$}
\\ \midrule
Without & \textbf{99.46 $\pm$0.03}  & \emph{92.15 $\pm$0.34} & \emph{82.23 $\pm$6.54} & \emph{99.11 $\pm$0.21} & \emph{63.50 $\pm$2.84} \\ \midrule
With & \emph{99.44 $\pm$0.02}  & \textbf{92.58 $\pm$0.22} & \textbf{83.96 $\pm$3.17} & \textbf{99.19$\pm$0.19} & \textbf{63.96$\pm$4.16} \\ 
\bottomrule
\end{tabular}
\end{tiny}
\end{center}
\label{pca_cosine_similarity_loss}
\end{table}

\section{Threat Analysis}

\begin{figure}[ht]
\begin{center}
\centerline{\includegraphics[width=7cm]{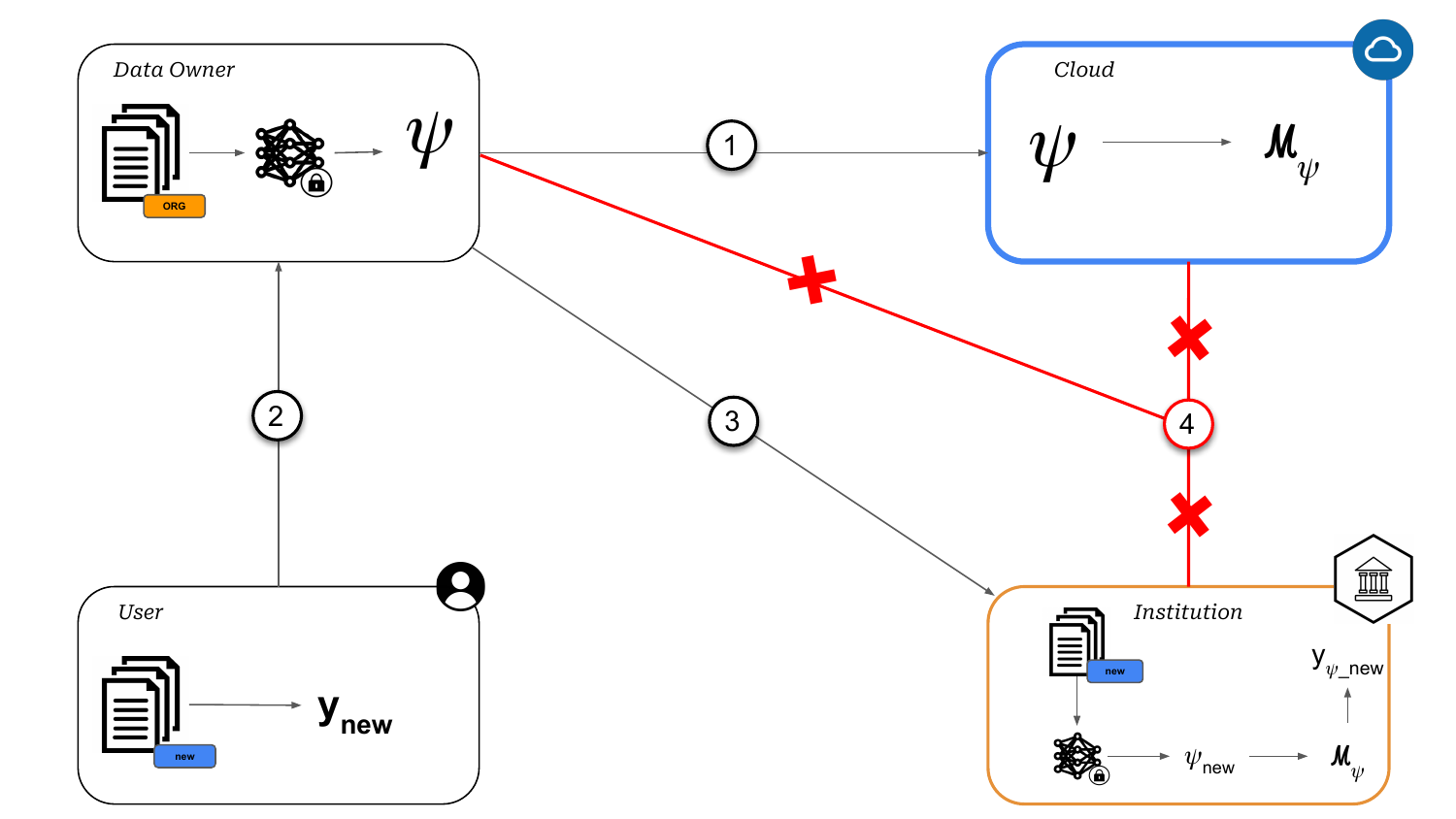}}
\caption{This figure summarizes the threat analysis schema with the four actors taking part in it. In $\textcircled{\tiny{\textbf{1}}}$ the adversary corrupts the cloud, in $\textcircled{\tiny{\textbf{2}}}$ the adversary corrupts a user, in $\textcircled{\tiny{\textbf{3}}}$ the adversary will have an access to one of the institutions and in $\textcircled{\tiny{\textbf{4}}}$ the adversary will corrupt both the cloud and one of the institutions. It is only in the latter scenario where the risk of inverting the original data might occur since the adversary will have access to the encoder $E(.)$ and the encoding of the original training samples \scalebox{1.2}{$\Psi$} thus a decoder $D_{inv}(.)$ can be trained on the pair of the institute data to perform an inversion attack on \scalebox{1.2}{$\Psi$}.}
\label{threat-analysis-scheme}
\end{center}
\end{figure}

In our solution, we consider four different actors: data owner(s), cloud server, users having access to inference services and institutions where encoder $E(.)$ and final model $\mathcal{M}_\Psi$ are shared. We assume that the cloud server, users, and institutions are honest but curious, meaning they are expected to follow the protocol. The main goal of the adversary corrupting these actors is to attempt to infer sensitive information about the original training samples by using their observations of the protocol execution.\\
The view of the adversary $\mathcal{A}$ corrupting the cloud consists of the encoded samples of the data owner(s) and the trained model on these samples. Since the adversary $\mathcal{A}$ has no knowledge about the utilized encoder by the data owner(s), it cannot return back to the original samples from the encoded samples. For instance, the adversary $\mathcal{A}$ knows neither the dimensionality nor the type of the original data. In addition to the protection of the privacy of the samples, our framework preserves the privacy of the model implicitly as well. Although the adversary $\mathcal{A}$ has access to the trained classifier model in plaintext, it has no use unless the adversary has access to the encoder. Therefore, we can conclude that our proposed framework securely allows the outsourcing of the computation to a third party or enables the third party to benefit from the output of the model without compromising the privacy of the data or the model.\\
The adversary $\mathcal{A}$ corrupting at least one user can perform predictions on the model $\mathcal{M}_\Psi$ for the encodings of given data. Through using the prediction service as an API , $\mathcal{A}$ has access only to the final predictions $y_{new}$. Without knowledge of the encodings of the training samples and the encoder $E(.)$, it is not possible to extract the training samples. $\mathcal{A}$ might still perform membership inference attacks which is a common problem for all machine learning models trained without differential privacy. Our solution does not aim to address these weaknesses.
The adversary, $\mathcal{A}$, who has compromised at least one institution, can access both the encoder $E(.)$ and the model $\mathcal{M}_\Psi$. $\mathcal{A}$ can use $E(.)$ to encode new data, and then use the encodings to make predictions using $\mathcal{M}_\Psi$, resulting in $(\Psi, y_{\Psi})$. Without the encodings of the original training samples, it is impossible to extract the training data. $\mathcal{A}$ can also perform attacks like model inversion and membership inference. However, as we mentioned, we do not present a solution to counteract these types of attacks in this article.\\
The adversary $\mathcal{A}$, who has compromised at least one institution and the cloud server, can access the encoder $E(.)$ and the encodings of the original training samples. In this scenario, $\mathcal{A}$ can train the decoder $D_{inv}(.)$ by using the data of compromised institution and the encoder $E(.)$ to reconstruct the original training samples. Therefore, in our security model, we assume that the cloud server and institutions do not collude. This assumption is practical in real-world scenarios. Data owners can conceal their identities from the cloud server, making it difficult to determine with whom the model trained by the cloud server is shared. The 4 scenarios described in our threat analysis are illustrated in Fig \ref{threat-analysis-scheme}.

\section{Limitations of the proposed framework}
Even though our encoding framework showed great performance-privacy results, it is for now still constrained to specific scenarios. Our framework is task-oriented which means that it requires the presence of a supervision task that guides our autoencoder for presenting a meaningful mapping. However, it is known within the ML community that most datasets do lack annotation \parencite{xu2020review,nguyen2021semi,humbert2022strategies} which currently excludes our proposed ppML encoding technique to be leveraged for unsupervised use cases. In addition, our model is not generic towards all types of data training distributions, in fact, our current framework is applicable only if the data is vertically split among input parties where they share the same sample ID but different feature spaces, like in the multimodal example used in our experiments, which excludes (for now) horizontal federated learning \parencite{yang2019federated}. 
 
\section{Conclusion \& Future Works}
In this study, we introduced an encoding strategy powered by representation learning leveraged for privacy purposes. The main goal and motivation behind implementing such a framework are to take advantage of discriminative representations learned in the hidden layers in the encoder part and take their concatenation as the encoding to be shared with other parties. To achieve this, we implemented a supervised residual autoencoder trained to consider both data reconstruction and an assigned classification task. To ensure a good representation we strengthened the training with two introduced losses one being the center loss applied on the concatenated encoding and a cosine similarity loss used to force the concatenation in having the same direction as the original input. We further presented the application workflow of our framework in a unimodal and multimodal settings. Our framework allows us to benefit from external computational resources  to perform training on the encoded data in a secure fashion since no information about the data is being revealed. As future work, we look forward to expanding the domain application of our encoding framework by including unlabeled datasets and horizontally distributed data for federated learning.  
\printbibliography
\newpage
\begin{appendices}
\section{Model Details}
\begin{figure*}[ht]
\begin{center}
\centerline{\includegraphics[width=\linewidth]{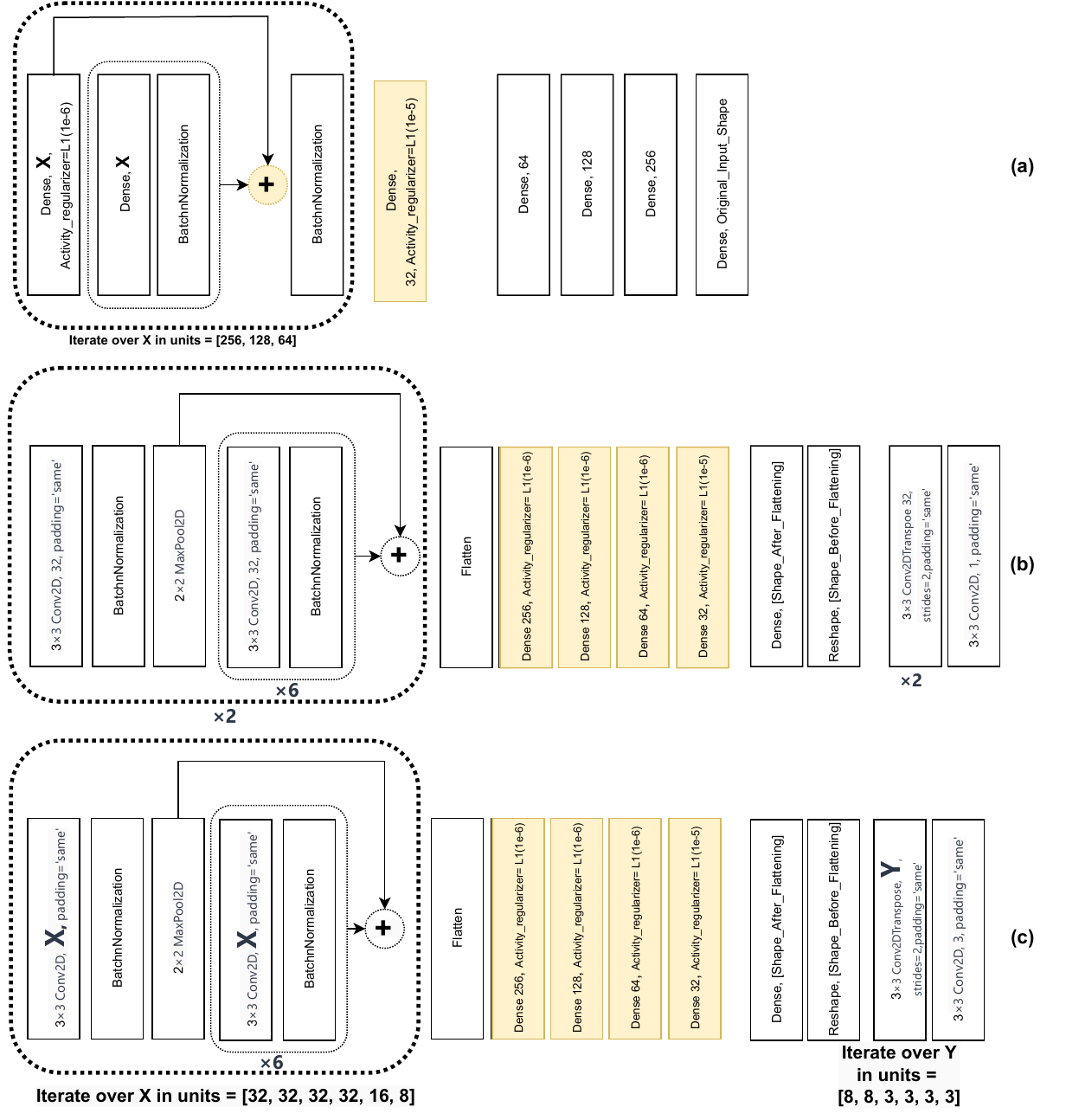}}
\caption{\textbf{(a)}, \textbf{(b)}, \textbf{(c)} are the model's backbone architecture for (Leukemia + TCGA), (MNIST+FashionMNIST) and (OCT) respectively. The layers colored with orange are the dense layers of the encoder part that are connected to a feed-forward neural classifier $f_{i}(.)$. As mentioned in Section 2.2, for better mapping invariance we applied $L1/L2$ kernel regularizer on all layers and $L1$ activity regularizer on the layer participating in the concatenated encoding (orange-colored). The final encoding \scalebox{1.2}{$\Psi$} however has the same shape $(N, 480)$ in the three versions.}
\label{architecture-types}
\end{center}
\end{figure*}

We present here more details about the type, number and hyperparameters of the layers composing the residual autoencoder in our framework. 
As previously mentioned in Section 3.1, two versions of supervised residual autoencoders were used in our experiments. If the modality is of type image then convolution layers  are included and we name it as C-SRAE. Otherwise we only include dense layers under the name SRAE. The latter is applied for the leukemia and TCGA datasets both being in a tabular format. C-SRAE is also presented in two different formats, a light version for small image input data as MNIST and FashionMNIST, and another for bigger image sizes as OCT with more convolution layers added. Fig \ref{architecture-types} presents more details about the three architectural types. However, for more uniformity the final encoding \scalebox{1.2}{$\Psi$} has the same shape $(N, 480)$ across the three presented versions. We also applied $L1/L2$ kernel regularization on every layer for better generalizability and $L1$ activity regularization on the dense layers taking part in the final concatenated encoding \scalebox{1.2}{$\Psi$} to ensure a proper invariant representation.   
We used \textit{Adam} as the optimizer and we monitored the training with an \textit{Early Stopping} callback on the validation loss.
\section{System Components}
Here we provide details about the hardware parts constituting our system.
\begin{table}[ht]
  \caption{Hardware Specifications}
  \label{system components}
  \centering
  \begin{tabular}{ll}
    \toprule
  
    Parameter     & Technical Specifications \\
    \midrule
    Memory & 32 GiB      \\
   Processor  & 11th Gen Intel® Core™ i7-11800H      \\
    No CPU Cores    & 16        \\
    CPU Frequency & 2.30 GHz \\
    GPU & Nvidia RTX A2000 Mobile \\
    \bottomrule
  \end{tabular}
\end{table}
\section{Training Details}
Here we provide more details about the set of hyperparameters related to the machine learning models used in our experiments and that were explored through a randomized grid search. The models and grid search were implemented using Sklearn library.  
\begin{table}[h]
 \small
  \caption{Hyperparameters for Randomized Grid Search}
  \label{grid_search}
  
  \centering
  \begin{tabular*}{\linewidth}{l|l|}
 
    \toprule
  
    ML Model & Hyperparameters \\
    \midrule
    KNN & \{"n\_neighbors": [i for i in range(2,10)]\}  \newline \\    \\
    \hline
    SVC & \{"C": [$10^i$ for i in range(-4, 2)]\} \newline \\    \\ \hline
    Decision Tree & \multicolumn{1}{p{10cm}}{\{"max\_depth": [2, 6],\newline "max\_depth":[10, 50, 100, 200, 300], \newline"min\_samples\_leaf": [1,3,5,10], \newline "criterion": ["gini", "entropy"], \newline "class\_weight":["balanced"]\}} \newline \\    \\ \hline
    Random Forest & \multicolumn{1}{p{10cm}}{ \{"n\_estimators":[100,200,300], \newline "max\_depth":[10, 50, 100, 200, 300] ,\newline "min\_samples\_leaf": [1,3,4,5,6,7],\newline "criterion": ["gini", "entropy"],\newline "class\_weight":["balanced"]\}} \newline \\    \\ \hline
    
    Multilayer Perceptron & \multicolumn{1}{p{10cm}}{\{"hidden\_layer\_sizes": [(10,30,10), \newline(20,), (100,)],\newline
        "activation": ["tanh", "relu"], \newline
        "solver": ["sgd", "adam"],
        \newline "alpha": [0.0001, 0.05], \newline
        "learning\_rate": ["constant","adaptive"]
    \}} \\
    \bottomrule
  \end{tabular*}
\end{table}
\end{appendices}
\end{document}